\documentclass[letterpaper, 10 pt, conference]{ieeeconf}  

\IEEEoverridecommandlockouts                              

\usepackage[noadjust]{cite}
\usepackage{graphicx}
\usepackage{algorithm}
\usepackage[noend]{algpseudocode}
\usepackage{svg}
\usepackage{amsmath}
\usepackage{multirow}
\usepackage{float}
\usepackage{subfig}

\title{\LARGE \bf
Towards Case-based Interpretability for Medical Federated Learning}

\author{Laura Latorre$^{1,2}$, Liliana Petrychenko$^{1,3}$, Regina Beets-Tan$^{1,3}$, Taisiya Kopytova$^{1}$, and Wilson Silva$^{1,4,5}$
\thanks{$^{1}$ Department of Radiology, The Netherlands Cancer Institute, Amsterdam, The Netherlands}%
\thanks{$^{2}$ Vrije Universiteit Amsterdam, Amsterdam, The Netherlands}%
\thanks{$^{3}$ GROW School for Oncology and Developmental Biology, Maastricht University Medical Center, Maastricht, The Netherlands}%
\thanks{$^{4}$ AI Technology for Life, Department of Information and Computing Sciences, Department of Biology, Utrecht University, Utrecht, The Netherlands}
\thanks{$^{5}$ Centre for Telecommunications and Multimedia, INESC TEC, Porto, Portugal}
}

\begin{document}

\maketitle
\thispagestyle{empty}
\pagestyle{empty}

\begin{abstract}
We explore deep generative models to generate case-based explanations in a medical federated learning setting. Explaining AI model decisions through case-based interpretability is paramount to increasing trust and allowing widespread adoption of AI in clinical practice. However, medical AI training paradigms are shifting towards federated learning settings in order to comply with data protection regulations. In a federated scenario, past data is inaccessible to the current user. Thus, we use a deep generative model to generate synthetic examples that protect privacy and explain decisions. Our proof-of-concept focuses on pleural effusion diagnosis and uses publicly available Chest X-ray data.
\end{abstract}


{\let\clearpage\relax
\section{INTRODUCTION}

Case-based interpretability is vital in explaining medical Artificial Intelligence (AI) model decisions. Generating explanations for AI model decisions is paramount to increasing trust and allowing widespread adoption in clinical practice~\cite{doshi2017towards}. We can find several approaches to producing explanations in the scientific literature, from saliency maps (highlighting image pixels driving the decision) to textual explanations~\cite{lipton2018mythos}.  However, in the context of radiology, case-based interpretability arises as a very natural interpretability strategy since it mimics the way radiologists would also explain or come up with a decision~\cite{silva2020interpretability}. When in doubt over a suspected disease condition, radiologists usually turn their attention to searching past cases and, based on disease-similarity to these past cases, reach a clinical decision.  Looking for adequate past cases is a time-consuming process that can be automated through the use of an AI retrieval model. AI retrieval models usually consist of finding a latent semantic representation of the original data (i.e., images) that is representative and adequate to measure the distance between cases and then evaluate disease similarity. All this process, though, starts from the premise that past data (which might have been used for training) is always available to the retrieval algorithm. 

Due to privacy concerns and regulations, sharing medical data is sometimes prohibitive, leading to a lack of heterogeneous training data and, consequently, to non-robust AI models that fail to generalize. It has been shown that even radiological images such as chest X-rays may contain biometric information that can be exploited by deep learning models to expose identity, therefore, infringing on data protection regulations~\cite{packhauser2022deep}. In order to overcome this limitation in data sharing, researchers have come up with federated learning approaches, which leverage all available data without the need to share it between institutions~\cite{pati2022federated}. This process works by distributing the model training to the institutional data owners and aggregating their results. Due to the privacy by default property of federated learning, it is conquering the fields of medical imaging and drug discovery~\cite{rieke2020future}. 

Even though federated learning’s potential to overcome some of the current AI flaws is currently widely recognized, it also introduces new challenges. The decentralized nature of federated learning guarantees compliance with privacy regulations but, at the same time, inhibits data access and inspection~\cite{augenstein2019generative}. Non-accessible data means that identifying bugs or detecting biases is impossible following conventional approaches. The same is true for case-based explainability. Since representative past cases may not be accessible for retrieval, as they might be stored in a different site, the whole concept of case-based interpretability loses its relevance, with consequences in the trust and understanding of AI classification model decisions. Moreover, the case-based interpretability role can be more extensive than just explaining decisions by also working as a computer-aided diagnosis.  In this work, we propose a proof-of-concept based on deep generative models to generate synthetic case-based explanations to be used in a medical federated learning setting.

\section{MATERIALS AND METHODS}

\subsection{Data}

Posterior-anterior (PA) chest radiographs from four large and publicly available databases (CheXpert~\cite{irvin2019chexpert}, MIMIC-CXR-JPG~\cite{johnson2019mimic}, BRAX~\cite{reis2022brax}, and VinDr-CXR~\cite{nguyen2022vindr}) were used. Since pleural effusion was present in all of the considered datasets, being one of the conditions with a more balanced distribution, it was chosen as the diagosis to perform and explain in our proof-of-concept. From each dataset, we only selected images labelled as having pleural effusion or not having pleural effusion, discarding all uncertain diagnoses. Based on the available data, we created an in-distribution test set (composed of 20\% of data from each of the local datasets: CheXpert, BRAX, and VinDr-CXR) and an out-of-distribution test set (composed of MIMIC-CXR-JPG data). In the federated learning setting, each dataset works as an independent hospital, i.e., as a local client.

\subsection{Method}

Our proposed methodology to obtain case-based explanations for federated learning models is presented in Fig.~\ref{fig:pipeline}. The first step requires training a privacy-preserving discriminative federated model used for diagnosis (in our case, pleural effusion prediction). The local discriminative models, and thus, the global model used, were based on the DenseNet-121 architecture~\cite{huang2017densely}. The global model is initialized with ImageNet~\cite{deng2009imagenet} pre-trained weights. In each round of the federated learning process, the central server distributes the current global model to the participating client devices (i.e., local sites corresponding to CheXpert, VinDR, and BRAX). Each client then independently trains the model using its local dataset. After each local training round, the global model is updated using federated averaging~\cite{mcmahan2017communication}. Differential privacy (DP) was implemented locally by using differential privacy optimizers (gradient noise injection techniques): DP-SGD and DP-Adam~\cite{abadi2016deep,mcmahan2018general}. By clipping and adding noise to the local client gradients, we avoid the risk of information leakage. The pseudo-code describing the training process in detail is presented in Algorithm 1. Our final model was first trained for 20 global epochs with frozen weights, updating just the last layer. Afterwards, the entire model was fine-tuned for 20 global epochs more. We considered 3 epochs for the local training process. Class weights corresponding to the inverse of the class frequency are used at each local site to mitigate the impact of class imbalance.        
\begin{algorithm}[h!]
\caption{\texttt{Federated Averaging}. The $K$ clients are
indexed by $k$; $B$ is the local minibatch size, $E$ is the number
of local epochs, and $\eta$ is the learning rate.} \label{fedlearn}
  \begin{algorithmic}
    \State \textbf{Data:} $(D_1, z_1), (D_2, z_2), ..., (D_k, z_k)$, 
    \State \quad \quad \quad $z_k \gets \frac{n_k}{\sum_k^K n_k}$: scaling factor; $n_k \gets$ \texttt{len(}$D_k$\texttt{)}: number of instances in $D_k$
    \State Initialize global model $w_0$
    \For{each round $t = 1, 2, ..., T$}
    \State \textbf{Local training process:}
      \For{each client $k = 1, 2, ..., K$}
        \State $w^k_{t} \gets$ \Call{ClientUpdate}{$k$, $w_{t-1}$}
      \EndFor
      \State \textbf{Model aggregating process:}
      \State $w_{t} \gets \sum_k z_k w_{t}^k$ (weighted average of resulting local models)
      \State \textbf{Local evaluation process:}
      \For{each client $k = 1, 2, ..., K$}
        \State Test the aggregated model $w_{t}$ using client validation set $D_{val}^k$
        \State $F1_t^k \gets$ \Call{F1Score}{$w_t$, $D_{val}^k$}
      \EndFor
      \State $F1_t \gets \sum_k z_k F1_t^k$
      \If{$F1_t>F1_{best}$}
        \State Save $w_t$ as best model
        \EndIf
    \EndFor
    \State
    \Procedure{ClientUpdate}{$k$, $w$}
        \State Receive global model $w$ and set its weights to local model  $w^{k}$
        \If{$t > t_{ft}$}
            \State Unfreeze $f$ layers of $w^{k}$
        \EndIf
        \For{each local epoch $i$ from 1 to E}
            \State Train $w^{k}$ with $B$ batch size and $\eta$ learning rate, using client dataset $D_{k}$
        \EndFor
        \State return $w^{k}$ to the server
    \EndProcedure
  \end{algorithmic}
\end{algorithm}

\begin{figure}[h!]
    \centering
\includegraphics[width=\linewidth]{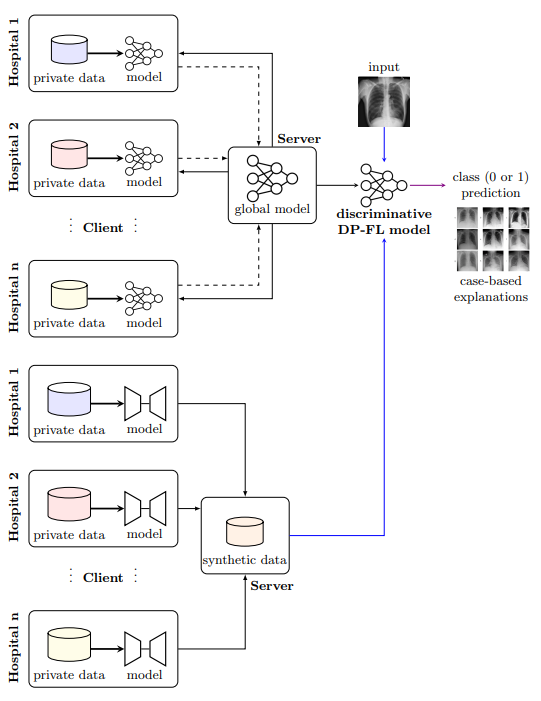}
    \caption{Case-based Explanations for Federated Learning.}
    \label{fig:pipeline}
\end{figure}

The second step of our methodology requires the training of a generative model to generate case-based explanations. In our experiments, we considered Medfusion~\cite{muller2022diffusion} as our generative model, given its previous promising results on Chest X-rays. A Medfusion model was trained for each of the clients, first training the Autoencoder, and later the Diffusion model. 400 samples with $t=150$ sampling steps, 200 with $label = 0$ and 200 with $label = 1$ were generated with each of the local models to create the synthetic dataset to be used to retrieve similar disease-matching cases as case-based explanations.

When a new case needs to be classified by the discriminative model, the three most similar synthetic samples for each of the clients are selected from the synthetic dataset to show them as case-based explanations. Similarity between images is computed based on the Normalized Euclidean distance in the feature space of the previous to the last layer of the model, which is formalized as:
\begin{equation}\label{normeucl}
    d_{DP-FL}(I_t,I_s) = \|F(\theta_{DP-FL},I_t)-F(\theta_{DP-FL},I_s)\|
\end{equation}
where $I_t$ represents the test image $t$, $I_s$ represents the synthetic image $s$, $\theta_{DP-FL}$ represents our model parameters, and $F$ represents the function that translates the original image into the latent representation. $d_{DP-FL}$ is computed for all the images from the synthetic dataset.


\subsection{Evaluation} 

\textbf{Baselines:} We considered a baseline for classification performance and a baseline for image retrieval. To provide an upper bound on the classification performance that could be achieved, we trained the same exact discriminative model but in a centralized setting (i.e., without the difficulties introduced by the different local distributions and differential privacy). As our retrieval baseline, we considered the standard structural similarity index measure (SSIM). The SSIM was computed between the query image and the retrieved case-based explanations. Since higher SSIM values represent higher similarity, images with higher SSIM were ranked first. 

\textbf{Metrics:} Due to the imbalance of the dataset, classification performance was measured by the F1-score. The quality of the retrieval was quantitatively evaluated by computing the normalised Discounted Cumulative Gain (nDCG) (Eq.~\ref{ndcg}) of the retrieval model ranking, compared to the ground truth ranking of the radiologist. $IDCG_p$ is the maximum possible value of the DCG metric, obtained when the ranking of the method is exactly the same as our ground truth. The subscript $p$ is the number of retrieved images considered for the evaluation (e.g., $p=9$). Finally, $rel_i$ represents the relevance value assigned to the ranking position $i$, with the least similar image having a relevance of 1 and the most similar image having a relevance of 5.

\begin{align}
    nDCG_p &= \frac{DCG_p}{IDCG_p}\label{ndcg}\\
    \text{where } DCG_p &= \sum_{i=1}^p \frac{2^{rel_i}-1}{log_2(i+1)}
\end{align}

\vspace{0.3cm}
\section{RESULTS}

We performed pleural effusion classification experiments considering a baseline centralized approach and the developed federated model. As expected, due to the distinct client distributions and inclusion of differential privacy, the federated learning model led to slightly worse results (Table~\ref{tab:confmatrixcentr}). We also generated saliency maps for both models by computing the SHAP values~\cite{lundberg2017unified}. As can be observed in Fig.~\ref{fig:shap}, when correctly predicting a pleural effusion case, both models highlighted the same regions.

\begin{table}[h!]
    \centering
    \caption{F1-score for the centralized model and the federated model in the in-distribution test set and out-of-distribution test set.}\label{tab:confmatrixcentr}
    \vspace{2mm}
    \begin{tabular}{l|l|c|c|c}
    \cline{3-4}
    \multicolumn{2}{c|}{}& In-distribution & Out-of-distr. &\multicolumn{1}{c}{}\\
    \cline{2-4}
    \multirow{2}{*}{}& Centralized Model & \textbf{0.857} & \textbf{0.864} & \\
    \cline{2-4}
    & Federated Model & $0.716$ & $0.781$ & \\
    \cline{2-4}
    \end{tabular}
\end{table}

\begin{figure}[h!]
\centering
\begin{minipage}[b]{.45\linewidth}
  \centering
  \centerline{\includegraphics[width=4.0cm]{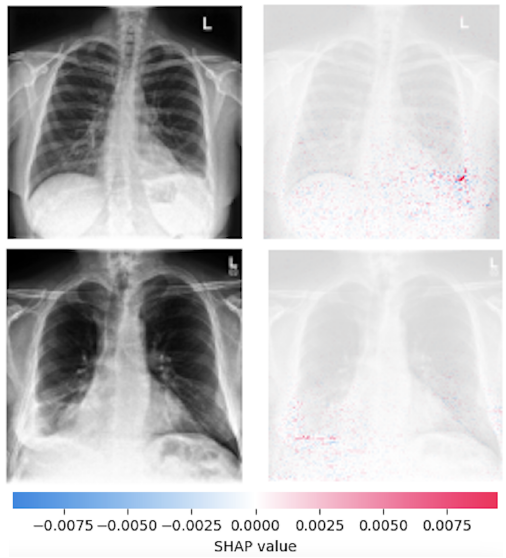}}
  \centerline{(a) Centralised Model}\medskip
\end{minipage}
\hfill
\begin{minipage}[b]{.45\linewidth}
  \centering
  \centerline{\includegraphics[width=4.0cm]{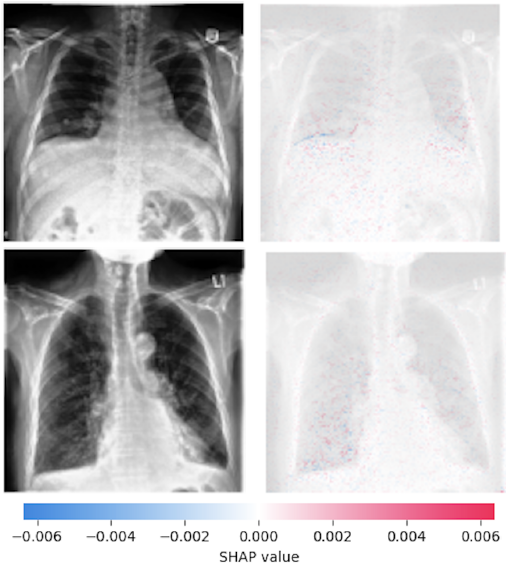}}
  \centerline{(b) Federated Model}\medskip
\end{minipage}
\caption{SHAP saliency maps for true positive test samples.}
\label{fig:shap}
\end{figure}

To evaluate the performance of the retrieval task, we considered 5 query images (1 negative and 4 positive) randomly chosen from the MIMIC test set. The pleural effusion prediction and 9 possible case-based explanations (3 from each client) for each query image were returned by our DP-FL model. Afterwards, the possible case-based explanations were assessed by a radiologist regarding two properties: realism and explanatory evidence. 
The radiologist was asked to label all the images, the queries and the case-based explanations, and assess their quality (or realism) for a proper pleural effusion diagnosis. Moreover, the radiologist also provided our ground truth by ranking the associated case-based explanations in relation to each of the test/query images, considering the similarity in terms of disease severity. During the qualitative evaluation by the radiologist, none of the samples was considered unreal or generated; the radiologist didn't realized the assessment was on AI generated images, and only considered that the quality was not very high, but comparable to those X-rays obtained in local hospitals. It was also emphasized that some images were cropped in a way that prevented a correct diagnosis of Pleural Effusion in one of the lungs, but this was also observed in two of the query images, indicating that this is a common property in the publicly available datasets, and thus our generative models are exposed to learn it. Fig.~\ref{fig:query-case} presents the Top-4 retrieved results for one of the test/query images. The ranking provided by our model and by SSIM is compared with the ground truth defined by the radiologist. As can be observed, the ranking provided by our model mimics better the ranking of a radiologist than the SSIM. The same was true for all the query images considered, which is shown by the retrieval scores presented in Table~\ref{tab:nDCG}.

\begin{figure}[h!]
    \centering
    \includegraphics[width=\linewidth]{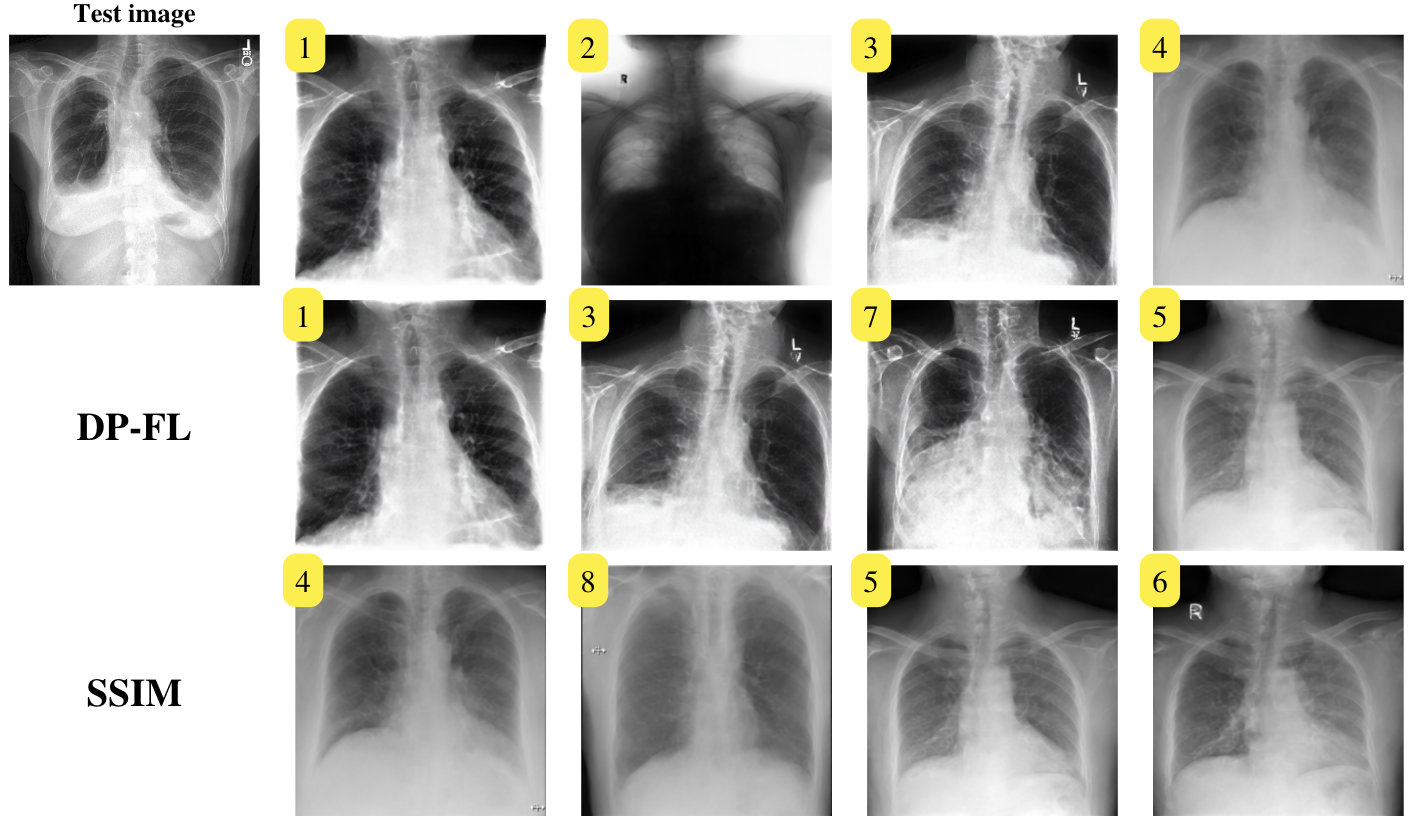}
    \caption{Example of a test case and the Top-4 retrieved explanations given by the radiologists (top), our model (DP-FL) and SSIM.}
    \label{fig:query-case}
\end{figure}

\begin{table}[h!]
    \caption{nDCG score for Model (normalized Euclidean Distance) and SSIM ranking.}\label{tab:nDCG}
    \vspace{2mm}
    \centering
    \begin{tabular}{c||c|c}
        Case & Proposed Model & SSIM \\
        \hline
        Test Image 1 & \textbf{0.921} & 0.588\\
        Test Image 2 & \textbf{0.854} & 0.645\\
        Test Image 3 & \textbf{0.765} & 0.693\\
        Test Image 4 & \textbf{0.931} & 0.610\\
        Test Image 5 & \textbf{0.901} & 0.617\\
    \end{tabular}
\end{table}

\section{DISCUSSION AND CONCLUSIONS}

Ensuring interpretability in the upcoming decentralized federated learning paradigms is paramount to increase trust and allow widespread AI adoption in clinical practice. In the context of radiology, case-based explainability is the most natural interpretability strategy. In this work, we investigated the use of a latent denoising diffusion probabilistic model in a federated chest X-ray classification task, demonstrating its potential to generate realistic and meaningful case-based explanations for a privacy-preserving federated learning discriminative model. While it is established how to retrieve case-based explanations in a standard centralized learning scenario, this is, to the best of our knowledge, the first study that addresses the challenge of generating case-based explanations in a privacy-preserving federated learning setting, where access to the entire data collection is restricted. However, to adopt the proposed method into a real-world medical application, a variety of further research is still needed. Future research should focus on exploring privacy-preserving generative models to guarantee that no identity information present in the client's data is leaked in the generated images. To be completely compliant with data protection regulations, we must ensure that the synthetic case-based explanations are not too similar to the sensitive training data nor expose identity. That could be done by either adding differential privacy or incorporating adversarial identity loss functions (e.g.,~\cite{montenegro2022disentangled}) into the training of the generative models.


}

\section*{ACKNOWLEDGMENT}
Research at the Netherlands Cancer Institute is supported by grants from the Dutch Cancer Society and the Dutch Ministry of Health, Welfare and Sport. This work is financed by National Funds through the FCT - Fundação para a Ciência e a Tecnologia, I.P. (Portuguese Foundation for Science and Technology) within the project CAGING, with reference 2022.10486.PTDC (DOI 10.54499/2022.10486.PTDC). The authors would like to acknowledge the Research High Performance Computing (RHPC) facility of the Netherlands Cancer Institute (NKI).




\bibliographystyle{IEEEtran}
\bibliography{references}

\end{document}